\pdfoutput=1

\documentclass[11pt]{article}

\usepackage[]{acl}

\usepackage{times}
\usepackage{latexsym}

\usepackage[T1]{fontenc}

\usepackage[utf8]{inputenc}

\usepackage{microtype}

\usepackage{inconsolata}
\usepackage{subfigure}
\usepackage{tabularx}
\usepackage{graphicx}
\usepackage{amsmath} 
\title{Connecting the Dots: Inferring Patent Phrase Similarity with Retrieved Phrase Graphs}

\author{Zhuoyi Peng \quad Yi Yang \\
        {The Hong Kong University of Science and Technology} \\
        \texttt{zpengaf@connect.ust.hk} \quad \texttt{imyiyang@ust.hk}}

\begin{document}
\maketitle
\begin{abstract}
We study the patent phrase similarity inference task, which measures the semantic similarity between two patent phrases. As patent documents employ legal and highly technical language, existing semantic textual similarity methods that use localized contextual information do not perform satisfactorily in inferring patent phrase similarity. To address this, we introduce a graph-augmented approach to amplify the global contextual information of the patent phrases. For each patent phrase, we construct a phrase graph that links to its focal patents and a list of patents that are either cited by or cite these focal patents. The augmented phrase embedding is then derived from combining its localized contextual embedding with its global embedding within the phrase graph. We further propose a self-supervised learning objective that capitalizes on the retrieved topology to refine both the contextualized embedding and the graph parameters in an end-to-end manner. Experimental results from a unique patent phrase similarity dataset demonstrate that our approach significantly enhances the representation of patent phrases, resulting in marked improvements in similarity inference in a self-supervised fashion. Substantial improvements are also observed in the supervised setting, underscoring the potential benefits of leveraging retrieved phrase graph augmentation.
\end{abstract}

\section{Introduction}
Patents are pivotal to the landscape of innovation, safeguarding novel ideas and fostering technological advancements \citep{hasan2010innovation}. Consequently, understanding patent phrase similarity becomes essential, offering insights into the nuances of intellectual property and aiding in the patent analytics applications, such as patent document categorization, patent retrieval, patent litigation analysis and so on \citep{tang2020multi, mase2005proposal, hall2007empirical}.  
For example, the United States Patent and Trademark Office (USPTO), in its mission to evaluate and grant patents, could leverage the patent phrase similarity task to streamline the examination process, identify prior art more efficiently, and ensure the distinctiveness of newly filed patent applications \citep{gao2022towards}.

Patent documents employ legal and highly technical language, featuring context-dependent terms that can deviate significantly from colloquial usage and may vary between different documents.
For instance, while a common term like "smartphone" might be easily recognized in everyday language, in patent documents it might be referred to as a "handheld electronic communication device", "mobile telecommunication apparatus", or even a "portable digital assistant with wireless capabilities".
Consequently, prevailing semantic textual similarity approaches, such as Sentence-BERT \citep{reimers-gurevych-2019-sentence} or SimCSE \citep{gao-etal-2021-simcse}, which focus on general text and rely on localized contextualized information for text representation, fall short in inferring patent phrase similarity. Additionally, obtaining a substantial collection of annotations from experts for supervised training presents significant challenges: the process is not only costly but also demands in-depth domain knowledge of the patent innovation landscape.
The intricate interplay between technically nuanced terms and the scarcity of labels renders the task especially challenging. A significant research gap persists: \textit{How can one effectively infer phrase similarity within the complex language of patents, especially in the absence of training labels}?


To address this challenge, we introduce a retrieval-based graph augmentation method to effectively capture phrase representations. Our approach is inspired by the strategies employed by real-world patent experts. For a given patent phrase, we begin by extracting a subgraph from the vast patent universe, such as those registered in the USPTO patent database. The derived subgraph comprises two node types: the phrase node and the patent node. A connection exists between a phrase node and a patent node if the phrase is present in the patent, and patent nodes interlink based on citation relationships. Hence, for a patent phrase such as "handheld electronic communication device", this subgraph might encompass related patents related to "portable telecommunication gadget". Concurrently, it could also surface interrelated phrases such as "wireless signal transceiver" and "handheld digital communicator". Such related patents and phrases provide a broader context, enabling a deeper understanding of the focal phrase. The information from its focal patent serves as the local context, whereas the extracted phrase graph offers a global context. This extracted phrase graph is subsequently processed through a graph attention network (GAT) \citep{velivckovic2018graph} to obtain its representative embedding. The final contextualized embedding of a patent phrase is a combination of its textual contextualized embedding and its associated phrase graph embedding. To address the issue of label scarcity, we utilize a self-supervised learning objective that capitalizes on the phrase graph's topology, facilitating the training of both the textual contextualized embedding and the graph learning parameters in an end-to-end manner.

We evaluate the proposed Retrieval Augmented Patent Phrase Similarity (RA-Sim) on a large patent phrase similarity dataset \cite{aslanyan2022patents}. In comparison with existing textual semantic similarity approaches—such as word embeddings (Word2vec \citep{mikolov2013efficient}, Glove \citep{pennington2014glove}, FastText \citep{bojanowski2017enriching}), contextualized embeddings (BERT \citep{devlin-etal-2019-bert}, RoBERTa \citep{liu2019roberta}), and semantic similarity embeddings (Sentence-BERT \citep{reimers-gurevych-2019-sentence}, Contriever \citep{izacard2022unsupervised}, SimCSE \citep{gao-etal-2021-simcse})—our RA-Sim method achieves substantial improvements on inferring patent phrase similarities in a self-supervised manner. Ablation studies and additional analyses further elucidate how RA-Sim enhances the patent phrase similarity inference task. Moreover, RA-Sim, when evaluated in supervised learning setting, consistently outperforms state-of-the-art methods. This highlights the potential advantages of incorporating retrieved graph for contextualized embedding augmentation. We release the codes for RA-Sim, enabling innovation and patent research scholars and practitioners to integrate it into their analytics pipeline.

\section{Related Work}
Our work is related to several lines of literature.

\textbf{Semantic Textual Similarity}. Semantic Textual Similarity (STS) is a classic natural language processing task. Word2Vec \citep{mikolov2013efficient}, FastText \cite{bojanowski2017enriching} train word-level embeddings unsupervisedly, which can be used for similarity inference. BERT \citep{devlin-etal-2019-bert} and RoBERTa \cite{liu2019roberta} pretrain language model, which provide powerful embeddings for similarity inference. Additionally, SBERT \citep{reimers-gurevych-2019-sentence} leverages dual-tower architecture to enhance sentence-level embeddings. SimCSE \citep{gao-etal-2021-simcse} proposes a self-supervised loss for similarity, by taking different views of the same sentence as contrastive postive \citep{jaiswal2020survey}. Usually, their embeddings are evaluated on Semantic Textual Similarity datasets \citep{agirre2012semeval, agirre2013sem, agirre2014semeval, agirre2015semeval, agirre2016semeval, cer2017semeval}. Recently, domain-specific semantic similarity tasks have drawn attention due to the potentially unique characteristics of domain language for similarity \citep{liu2024surface}. Patent phrase similarity inference also imposes some unique challenges. Firstly, methods that perform well for general text, such as SBERT and SimCSE, and even domain-specific model Patent-BERT \citep{srebrovic2020leveraging}, perform poorly due to the interplay of technical language and phrase brevity, implying a need for domain-specific modelling. Moreover, labelling patent phrase similarity requires a high expertise in understanding the patent landscape, which is often lacking in practice, resulting in label absence for model training.
 
\textbf{Retrieval-based NLP Generation}. A growing body of research incorporates a retrieval system for NLP generation tasks \citep{asai2023tutorial, yogatama2021adaptive, borgeaud2022improving, zhong-etal-2022-training,tang2024multihop}. Specific applications include question answering \citep{kumar2016ask, de2019episodic, chen-etal-2023-augmenting}, dialogue \citep{fan2021augmenting} and other traditional NLP
tasks \citep{lewis2020retrieval}. No prior work has utilized a retrieval module for the tasks in patent domain.

\textbf{Graph Learning for Patent Analysis}. Graph neural networks \citep{kipf2017semisupervised, hamilton2017inductive, velivckovic2018graph} have been used for analyzing patents \citep{tang2020multi, fang2021patent2vec, siddharth2022engineering, zuo2022patent} or other technical text like customer requirements \citep{shbita2023understanding}. The potential of graph is still underexplored in patent phrase similarity inference.

This paper presents the first self-supervised framework for patent phrase similarity inference, by retrieving a domain graph to amplify the global contextual information for patent phrases.

\begin{figure*}[h]
    \centering
    \includegraphics[width=16cm]{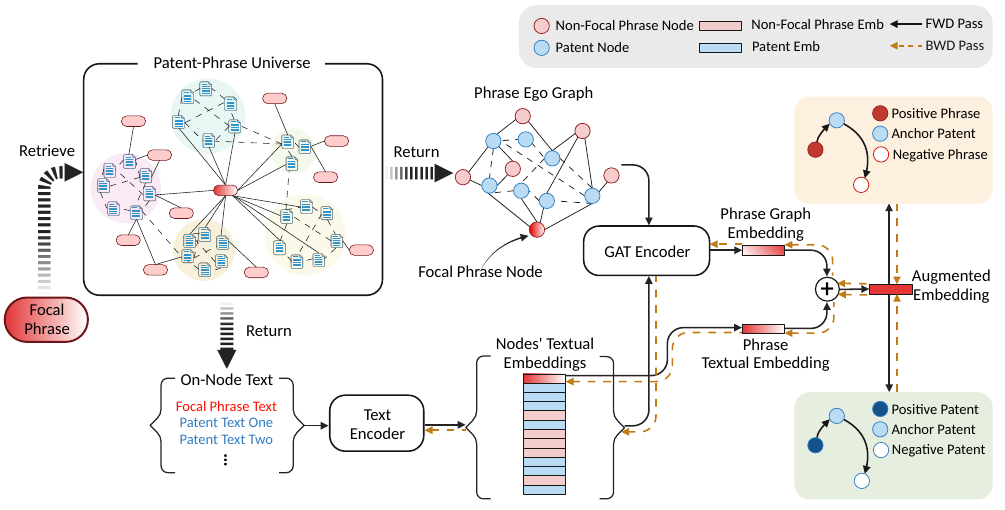}
    \caption{Design framework of RA-Sim: An ego graph for a phrase is retrieved from the patent-phrase universe to complement the contextual information for a patent phrase. Two self-supervised objectives, namely the retrieval contrastive loss and the citation contrastive loss, are employed to train the text and the graph encoders jointly.}
    \label{fig:framework}
\end{figure*}
\section{Retrieval Augmented Patent Phrase Similarity}

Patent phrases typically employ legal and highly technical language and may deviate significantly from colloquial usage. 
Moreover, patent phrases are typically short, containing only a few tokens. This brevity makes it even more challenging to meaningfully represent them due to the lack of contextual information.
For example, consider the technical phrases "acid absorption" and "chemically soaked" from the Patent Phrase Similarity dataset released by Google \citep{aslanyan2022patents}. Even though these two phrases don't share any words, patent experts rate them as domain-related, assigning a similarity score of 0.25 on a scale of [0,1]. Conversely, the phrases "acid absorption" and "acid reflux" are rated as "not related" with a similarity score of 0\footnote{This task is challenging, as evidenced by GPT-4's ratings: it deems "acid absorption" and "acid reflux" to be more similar than "acid absorption" and "chemically soaked".}. 

\noindent\textbf{Design Rationale}: To enhance patent phrase similarity inference, we propose to integrate additional contextual information to bolster phrase representations. There are two primary avenues to augment this contextual information. First, the patents in which a phrase appears can offer crucial context to elucidate the phrase's meaning. Second, the citation network linking the patents can further refine the representations of these patents, which, in turn, enriches the contextual understanding of the associated phrases. Building on this premise, we introduce a retrieval-augmented patent phrase similarity method, termed \textbf{RA-Sim}. The design framework of RA-Sim is illustrated in Figure \ref{fig:framework}.

\subsection{Constructing Patent-Phrase Universe}
We begin with a patent database that includes granted patents along with their citation information. In this study, our emphasis is on the U.S. granted patent data from USPTO. It's worth noting that while our focus is on U.S. patents, RA-Sim is not exclusive to them and can be generalized to patents granted in other jurisdictions as well.

Formally, we have a set of $N$ patents $\mathcal{V}$. For the $i$-th patent $v_i \in \mathcal{V}$, its text (e.g., patent abstract) is denoted as $d_i$. We represent $\mathcal{E}^c \in \mathcal{R}^{N \times N}$ as the adjacency matrix indicating citation relations between patents. Regarding the citation relation, $\mathcal{E}_{i j}^c=1$ if patent $v_i$ cites patent $v_j$, or 0 otherwise. In our context, the USPTO patent dataset contains 7,619,250 utility patents with 102,674,056 citations.

The original patent database does not include a specific list of phrases associated with each patent. Given our emphasis on patent phrase similarity, we augment the patent citation graph with phrase nodes. For the sake of efficiency, we employ the Rapid Automatic Keyword Extraction (RAKE) algorithm \citep{rose2010automatic} to extract key phrases from the patent set $\mathcal{V}$. This process yields a set of $M$ phrase $\mathcal{U}$. We present some phrase examples in \textbf{Appendix} \ref{appendix_phrase}. 

After obtaining the patent set $\mathcal{V}$ and phrase set $\mathcal{U}$, we establish relations between these two sets. Specifically, for a given patent phrase $u \in \mathcal{U}$, we utilize a retrieval algorithm to fetch the top-
$k$ patents from the patent set $\mathcal{V}$. Following prior work \citep{zhong-etal-2022-training}, we adopt BM25 as the retrieval algorithm, due to efficiency and capability\footnote{We also examine different retrieval systems like Doc2Vec \citep{le2014distributed} and Contriever \citep{izacard2022unsupervised}, among which BM25 performs the best.}.
We denote $\mathcal{E}^r \in \mathcal{R}^{N \times M}$ as the adjacency matrix, indicating the relationships where a patent is retrieved based on a patent phrase query. The retrieval examples are shown in \textbf{Appendix} \ref{appendix_retrieval}. 

In the patent-phrase universe, there are two types of nodes: the phrase node and the patent node. A connection is established between a phrase node and a patent node if the phrase appears in the patent. Additionally, patent nodes are interconnected based on their citation relationships.
These associated patents and phrases offer a broader context, enabling a deeper understanding of the focal phrase. The patent phrase universe  is denoted as $\mathcal{G}=\left(\mathcal{U}, \mathcal{V}, \mathcal{D}, \mathcal{E}^r, \mathcal{E}^c\right)$.

\subsection{Phrase Representation Augmentation with Phrase Ego Graph}
The semantic information of a patent phrase retains a wealth of knowledge pertinent to phrase similarity inference. Without loss of generality, let's denote a textual encoder $f$ that maps a patent phrase $u$ to a $d$-dimensional numerical vector $h_u^f$, i.e.,  $f(u) \mapsto h_u^f \in \mathcal{R}^d, \forall u \in \mathcal{U}$. For instance, one can choose BERT \citep{devlin-etal-2019-bert} or RoBERTa \citep{liu2019roberta} as the textual encoder.  

Furthermore, for a patent phrase $u \in \mathcal{U}$, we retrieve its corresponding ego graph from the patent phrase universe, referred as phrase ego graph, $\mathcal{G}_u=\left(\mathcal{U}_u, \mathcal{V}_u, \mathcal{E}^r_u, \mathcal{E}^c_u\right)$, using a recursive neighbor sampling approach. Specifically, to generate the ego graph for the patent phrase $u$, we follow these steps: 1). Initialize a node set that includes only $u$ and an empty edge set; 2).
Sample neighbors for each node within the node set. Add the resulting neighbor nodes to the node set and their corresponding neighboring edges to the edge set;
3). Repeat step 2 for a predetermined number of iterations. 
By iteratively expanding both the node set and edge set, we construct a patent phrase ego graph, $\mathcal{G}_u$, comprising patents where the focal phrase $u$ is present, along with a set of related patents referenced within those patents.

We subsequently employ a graph neural network $g$ to transform $\mathcal{G}_u$ to a $d$-dimensional vector $g(\mathcal{G}_u)$. In essence, 
$g$ functions as a readout operation in graph neural networks \citep{kipf2017semisupervised, hamilton2017inductive, velivckovic2018graph}, i.e., $g(\mathcal{G}_u) \mapsto h_u^g \in \mathcal{R}^d$. 
This fixed-size representation $h_u^g$ encodes the ego-graph information of focal patent phrase $u$, which provides an augmentation to its textual representation. 

Finally, the retrieval-augmented phrase embedding for $\phi(u)$  is modelled as follows: 
\begin{equation}\label{eq:context_embedding}
    \phi(u) = f(u) \oplus g(\mathcal{G}_u),
\end{equation}
where $\oplus$ is  element-wise addition.  
For two given patent phrases of interest, the similarity between them is measured by the cosine similarity of corresponding augmented embeddings.


\subsection{Phrase Ego Graph Representation}
In this section, we discuss how to map a phrase $u$'s ego graph $\mathcal{G}_u$ to a fixed-size representation $g(\mathcal{G}_u)$.

\noindent\textbf{Initialize Node Embeddings by Textual Encoder}. Recall that the nodes in $\mathcal{G}^{u}$ are either patent phrases or patent texts. Therefore, we reuse the text encoder $f$ to map the patent phrase node  $i \in \mathcal{U}_u$ and patent node $j \in \mathcal{V}_u$ in $\mathcal{G}_u$ to its initial embeddings:

\begin{equation}
    \boldsymbol{h}(i)^0=f(i), \forall i \in \mathcal{U}_u, {~~~} \boldsymbol{h}(j)^0=f(j), \forall j \in \mathcal{V}_u.
\end{equation}

\noindent\textbf{Graph-based Transformation}.
We then use Graph Attention Network (GAT) \citep{velivckovic2018graph} to recursively performing feature transformation on the ego graph $\mathcal{G}^{u}$. Detailed GAT transformation can be found in \citep{shi2020masked}. Note that the graph-based transformation is architecture-agnostic and can use other GNN architectures as well. 
We use GAT to model the two relations in the ego graph, including retrieval relation and citation relation, respectively in $l$-the layer, as follows. 

For a given phrase node $i$ in ego graph, we aggregate neighbors features associating with the $i$'s feature at $l$th-layer:

\begin{equation}
    \boldsymbol{h}(i)^{l+1}=\operatorname{GAT}^{l}\left(\boldsymbol{h}(i)^l,\left\{\boldsymbol{h}(j)^l\right\}_{j \in \mathcal{N}^r_u(i) }\right),
\end{equation}
where $\mathcal{N}^r_u(\cdot)$ returns the retrieved patents for phrase $i$ in its ego graph $\mathcal{G}_u$.

For a given patent node $j$ in ego graph, we model two types of feature aggregation from retrieval and citation perspectives by:
\begin{equation}
\boldsymbol{h}(j)_{\text {retrieval}}^{l+1}=\operatorname{GAT}^{l}_{r}\left(\boldsymbol{h}(j)^l,\left\{\boldsymbol{h}(i)^l\right\}_{i \in \mathcal{N}^r_u(j) }\right),
\end{equation}
and 
\begin{equation}
\boldsymbol{h}(j)_{\text {citation}}^{l+1}=\operatorname{GAT}^{l}_{c}\left(\boldsymbol{h}(j)^l,\left\{\boldsymbol{h}(i)^l\right\}_{i \in \mathcal{N}^c_u(j) }\right).
\end{equation}
where $\mathcal{N}^r_u(\cdot)$ and $\mathcal{N}^c_u(\cdot)$ are neighbouring lookup functions to return phrase and patent neighbors for patent $j$ in ego graph $\mathcal{G}_u$, respectively.

Then, we combine the above two feature transformations to form patent node $j$'s representation at the $l+1$ layer as follows: 
\begin{equation}
    \boldsymbol{h}(j)^{l+1}=\operatorname{Mean}\left(\boldsymbol{h}(j)_{\text {retrieval}}^{l+1}, \boldsymbol{h}(j)_{\text {citation}}^{l+1}\right),
\end{equation}
where $\operatorname{Mean}(\cdot,\cdot)$ is element-wise Mean pooling.


Finally, the retrieval-augmented phrase embedding $\phi(u)$ in Equation \ref{eq:context_embedding}, can be written as follows:
\begin{equation}\label{eq:context_embedding_final}
    \phi(u) = f(u) \oplus  g(\mathcal{G}_u)=\boldsymbol{h}^{0}(u)\oplus \boldsymbol{h}^{l}(u),
\end{equation}
where $l$ is the number of GAT layers.

\subsection{Learning Objective}
One challenge in inferring patent phrase similarity is the absence of annotated similarity labels. To address this issue, we propose training both the textual encoder $f$ and the graph encoder $g$ using a self-supervised learning objective. 

First, we posit that within the ego graph, the representation of a given patent node should be more similar to the phrase node that retrieves the patent than to a random phrase node unlinked to that patent.
We adopt the following triplet margin loss to capture this graph topology:
\begin{equation} \label{eq:loss_retrieval}
\begin{aligned}
\mathcal{L}_{retrieval}(a, p, n)=&\max \{\operatorname{dis}(\boldsymbol{h}(a), \boldsymbol{h}(p)) \\
&-\operatorname{dis}(\boldsymbol{h}(a), \boldsymbol{h}(n))+\delta_r, 0\},
\end{aligned}
\end{equation} 
where $a \in \mathcal{V}_u$ denotes a patent node (anchor), and $p \in \mathcal{U}_u$ denotes a phrase node (positive) and $p \in \mathcal{U}$ denotes an in-batch negative phrase (negative).  $\delta_r$ is a hyperparameter denoting the predefined margin. $\operatorname{dis}(\cdot, \cdot)$ is the Euclidean distance function.

Second, the representation of a given patent node should be more similar to another patent node with which it shares a citation, compared to a random patent node that has no linkage to that patent.
We capture the citation connections by:
\begin{equation}\label{eq:loss_citation}
\begin{aligned}
\mathcal{L}_{citation}(a, p, n)=&\max \{\operatorname{dis}(\boldsymbol{h}(a), \boldsymbol{h}(p))  \\
&-\operatorname{dis}(\boldsymbol{h}(a), \boldsymbol{h}(n))+\delta_c, 0\},
\end{aligned}
\end{equation} 
where $a, p \in \mathcal{V}_u$ denote two patents with a citation relation, and $n \in \mathcal{V}$ is an in-batch negative patent. $\delta_c$ is the margin hyperparameter. 

Our final learning objective is shown as follows (the notations for sampled positive $p$ and negative $n$ are dropped for simplicity and readability):
\begin{equation}\label{eq:final_loss}
    \mathcal{L}=\alpha \sum_{u \in \mathcal{U}_u} \mathcal{L}_{\text {retrieval}}(u) + (1-\alpha) \sum_{v \in \mathcal{V}_u} \mathcal{L}_{\text {citation}}(v),
\end{equation} 
where $\alpha$ is a coefficient to balance the importance between two optimization goals. Note that the framework is trained in a end-to-end fashion where the parameters in $f$ and $g$ are jointly optimized.


\section{Experiment}
We evaluate proposed RA-Sim method on Patent Phrase Similarity Dataset \citep{aslanyan2022patents}, which is curated by Google and rated by domain experts in patents . 





\subsection{Data}

\textbf{Patent Phrase Similarity} \citep{aslanyan2022patents}. The dataset comprises annotated patent phrase pairs, including 36,473 for training, 2,843 for validation, and 9,232 for testing. Given that our method employs self-supervised learning and does not utilize annotated patent similarities, we restrict our evaluation to the testing set. Later in the experiments, we will explore a supervised setting for RA-Sim, where labeled training data are utilized.



As highlighted previously, an ego phrase graph is retrieved from the patent-phrase universe, which is constructed by the following two external datasets.

\textbf{USPTO Patent Dataset}. We download the data of US granted patents and their citations from the PatentsView\footnote{https://patentsview.org/download/data-download-tables}. This dataset encompasses approximately seven million patents issued from January 6, 1976, to December 28, 2022. From this collection, our sample includes 7,619,250 utility patents. Additionally, we retrieve 102,674,056 citation records. Patent abstracts are extracted from the database to serve as the source of patent textual information. 

\textbf{RAKE Phrase Set}. We also create a patent phrase dataset using RAKE algorithm \cite{rose2010automatic}, for efficient phrase generation for the large patent database. For each patent abstract, we extract 3 key phrases to obtain 22,852,178 phrases and remove simple digits or single alphabets. We also apply WordNet Lemmatizer\footnote{https://www.nltk.org/\_modules/nltk/stem/wordnet.html} to lemmatize phrases. Then we remove the phrases whose frequency is less than 25 which drops a large amount of rare phrases. This process results in a phrase set of size 26,555. We summarize the token statistics in \textbf{Appendix} \ref{APP:tokenstatistics}.

\subsection{Metrics}
Following prior research \citep{aslanyan2022patents}, we use two metrics to evaluate similarity inference performance: Pearson correlation and Spearman correlation which measure the relatedness between inferred and labelled similarities. A higher score indicates better alignment with patent experts and thus better inference performance. The reported results are averaged over 3 runs.

\subsection{Training Setup}
Our model is trained on 4 RTX 3090 GPUs. We use Sentence-BERT model with all-mpnet-base-v2\footnote{We examine model performance under different textual encoders like GTE \citep{li2023towards} and E5 \citep{wang2022text}, and all-mpnet-base-v2 performs the best.} as the textual encoder. The full training details can be found in the \textbf{Appendix} \ref{app:fulltrainingdetails}.

\subsection{Baselines}

We compare RA-Sim with existing methods and two proposed baselines. The full details of the baselines are provided in the \textbf{Appendix} \ref{app:baselinemodels}.

The pretrained baselines include base/large \textbf{BERT} \citep{devlin-etal-2019-bert}, base/large RoBERTa \citep{liu2019roberta}, Sentence-BERT \citep{reimers-gurevych-2019-sentence} (\textbf{SBERT}), and pretrained \textbf{Contriever} \citep{izacard2022unsupervised} for dense information retrieval. We use mean pooling to obtain phrase-level embeddings. We also evaluate \textbf{Patent-BERT}, which is pretrained on patent data, for comparison. Moreover, we compare multi-stage contrastive embedding method \textbf{GTE} \citep{li2023towards} and the instruction-finetuned model \textbf{instructor-xl} \citep{su2022one} with our method.

We fine tune base/large BERT, base/large RoBERTa and SBERT with contrastive loss SimCSE \citep{gao-etal-2021-simcse}, and derive baselines \textbf{$\text{SimCSE-BERT}_\text{base}$}, \textbf{$\text{SimCSE-BERT}_\text{large}$}, \textbf{$\text{SimCSE-RoBERTa}_\text{base}$}, \textbf{$\text{SimCSE-RoBERTa}_\text{large}$} and \textbf{SimCSE-SBERT}.

We also compare established off-the-shelf word-level embeddings, including \textbf{Glove} \citep{pennington2014glove}, \textbf{Word2Vec} \citep{mikolov2013efficient}, and \textbf{FastText} \citep{bojanowski2017enriching}.

We propose a retrieval baseline and a graph baseline: \textbf{RetrieveAvg} and \textbf{Graph-Only}. RetrieveAvg is a retrieval-based method that retrieves the most relevant patent for a given phrase using BM25 and obtains an augmented embedding by weighted averaging the phrase embedding and the most relevant patent's embedding. Graph-Only is a graph-based baseline that replaces the initial GAT embeddings in RA-Sim with random embeddings, to remove the effects of phrase and patent text information.

\subsection{Main Results}
We show the main results of experiments in Table \ref{table:main_results}, leading to the following observations.

\begin{table}[h]

\small
\begin{tabular}{lccc}
\hline Model & Dim. & Pear. Cor. & Spear. Cor. \\
\hline 

GloVe$\dagger$ & 300 & 0.429 & 0.444 \\
FastText$\dagger$ & 300 & 0.402 & 0.467 \\
Word2Vec$\dagger$ & 250 & 0.437 & 0.483 \\
Patent-BERT$\dagger$ & 1024 & 0.528 & 0.535 \\

Contriever & 768 & 0.528 & 0.498 \\

$\text{BERT}_{\text{base}}$ & 768 & 0.413 & 0.418 \\

$\text{BERT}_{\text{large}}$ & 1024 &  0.422 & 0.405 \\

$\text{RoBERTa}_{\text{base}}$ & 768 & 0.313 & 0.329\\

$\text{RoBERTa}_{\text{large}}$ & 1024 &  0.364 &  0.372 \\

$\text {SimCSE-BERT}_{\text {base}}$ & 768 &0.525 &0.516 \\

$\text {SimCSE-RoBERTa}_{\text {base}}$ & 768 &0.471 &0.435 \\

$\text {SimCSE-BERT}_{\text {large}}$ & 1024 &0.534 &0.510 \\

$\text {SimCSE-RoBERTa}_{\text {large}}$ & 1024 &0.484 &0.460 \\

$\text {SimCSE-SBERT}$ & 768 &0.562 &0.542 \\

$\text {GTE}_{\text {base}}$ & 768 & 0.586 & 0.562 \\

$\text {GTE}_{\text {large}}$ & 1024 & 0.599 & 0.573 \\

Instructor-xl & 768 & 0.600 & 0.584 \\

SBERT & 768 & 0.598 & 0.577 \\

Graph-Only & 768 & 0.258 & 0.146 \\

RetrieveAvg & 768 & 0.622 & 0.595 \\

\textbf{RA-Sim} & 768 & \textbf{0.633} & \textbf{0.629}
\\
\hline
\end{tabular} 
\caption{Patent phrase similarity inference performance under self-supervised setup, in terms of Pearson Correlation and Spearman Correlation. $\dagger$ denotes the scores reported by previous work \cite{aslanyan2022patents}. Results are averaged over 3 runs.}
\label{table:main_results}

\end{table}

\textbf{RA-Sim outperforms existing state-of-the-art methods}. Our method leverages a phrase ego graph to augment embeddings and achieves a significant improvement over the second-best model, SBERT, with a 5.8\% increase in Pearson R and a 9\% gain in Spearman R. Notably, the state-of-the-art method in general text similarity, SimCSE, exhibits poor performance with BERT, RoBERTa, and SBERT as base models, with a decline from 0.598 to 0.562 in Pearson R in SimCSE (SBERT). This suggests that SimCSE is not an appropriate fine-tuning objective for inferring patent phrase similarity, highlighting a need for self-supervised objective in patent phrase similarity task. Moreover, we find that the multi-stage contrastive learning method GTE and the instruction-finetuned model instructor-xl do not significantly outperform SBERT, while our RA-Sim outperforms them in a large margin. This implies that improving model capability in general text does not necessarily improve the performance in patent phrase similarity inference. Lastly, RA-Sim beats the state-of-the-art method in dense information retrieval (\textit{i.e.}, Contriever) and a series of word embedding methods (\textit{e.g.}, Word2Vec) by a large margin, demonstrating the superiority of RA-Sim.

\textbf{Retrieval module is helpful}. Retrieval-based methods, including our method and the proposed RetrieveAvg baseline, underscore the effectiveness of retrieval in improving patent phrase similarity inference. For example, our proposed baseline RetrieveAvg can improve SBERT from 0.598 to 0.622 in Pearson R. These results indicate that the retrieval module is of vital importance.

\textbf{Phrase graph is informative}. 
From the result of Graph-Only, we observe that incorporating structural information (i.e., relations) alone contributes to patent phrase similarity inference to a certain extent, leading to a Pearson R of 0.258. This observation highlights the informative role of graph topology in augmenting phrase embeddings. 


\textbf{Domain patent texts matter}. The patent document text information from USPTO is indispensable for RA-Sim, as demonstrated by the large performance degradation of the Graph-Only baseline due to lacking domain text information, only presenting a Spearman Correlation of 0.146.



\subsection{Ablations}

A series of ablation results are shown in Table \ref{table:ablation}.

\textbf{Phrase Number and RAKE Bias}. We find that model performance drops from 0.633 to 0.616 in Pearson R when only using 1\% phrases. However, RA-Sim only experiences a minor performance degradation with 10\% phrases, which validates its robustness. One may also concern about the bias introduced by RAKE. When we replace the phrase set with the phrases from the training similarity dataset (keeping the training labels absent), we observe no significant change in terms of performance. When dropping lemmatization, the performance remains. Thus, our constructed RAKE phrase set is capable for effectively training RA-Sim.

\begin{table}[]
\small

\begin{tabular}{lcc}
\hline Ablation & Pear. Cor. & Spear. Cor. \\ \hline
1\% RAKE Phrases & 0.616 & 0.613 \\
10\% RAKE Phrases & 0.633 & 0.621 \\
$k$=3 in BM25 & 0.626 & 0.621 \\
$k$=7 in BM25 & 0.629 & 0.626 \\
$k$=50 in BM25 & 0.621 & 0.624 \\
Expansion Iter=3 & 0.620 & 0.619 \\
$\mathcal{L}_{citation}$ Only & 0.551 & 0.547 \\
$\mathcal{L}_{retrieval}$ Only & 0.608 & 0.588 \\
Hard Negatives & 0.602 & 0.582 \\
Additive Attention & 0.613 & 0.604 \\
GCN & 0.605 & 0.601 \\
GCNII & 0.612 & 0.615 \\
w/o Lemmatization & 0.629 & 0.624 \\
Use Train Set Phrases  & 0.635 & 0.631 \\
\textbf{RA-Sim ($k$=5 in BM25)}  & \textbf{0.633} & \textbf{0.629} \\
\hline
\end{tabular}\caption{Ablation results.} \label{table:ablation}

\end{table}

\textbf{Retrieval Size in BM25}. We evaluate our method with different BM25 retrieval sizes (best performance is achieved when $k$=5). The results imply that: the receptive field size of the phrase ego graph (controlled by $k$) should be neither too small nor too large. If the retrieved ego graph is too small, it may not provide enough information to enhance the embedding. Conversely, if the $k$ in BM25 is too large, the retrieved two graphs may become too blurry to distinguish two different phrases.

\textbf{Iteration Time for Expanding Graph}. We change the iteration time for expanding the phrase graph to three iterations. This change results in a decrease in performance compared to the setup with two iterations used in the main result setup. It appears that increasing the iteration time leads to the inclusion of irrelevant or noisy information, thereby diluting the power of the constructed phrase graphs.

\textbf{Hard Negatives}. One may consider a more advanced negative scheme. We replace the negatives in Eq. \ref{eq:loss_retrieval} and \ref{eq:loss_citation} with the structure-aware negatives in prior work \cite{ahrabian-etal-2020-structure}. The intuition behind is to utilize the ego graph structure to generate hard negatives from the anchor's $n$-hop neighborhood. However, when using 3-hop negatives, we find a performance degradation. We suspect that: in an ego graph, the neighbors within a 3-hop distance and the anchor node itself are highly relevant, and should not be treated as hard negatives.

\textbf{Graph Layer}. We also implement the GAT layer with Additive Attention \citep{velivckovic2018graph} scheme or use GCN \citep{kipf2017semisupervised} or GCNII \citep{chen2020simple}. However, the graph layer with the attention scheme we used achieves the best empirical performance, consistent with prior findings \citep{shi2020masked}.

\textbf{Effectiveness of Two Losses}.
Table \ref{table:ablation} shows that there is only a minor improvement when we solely use retrieval loss. This illustrates the importance of citation information. When solely using citation loss, the performance of the model degrades largely in Pearson R (from 0.633 to 0.551). This reveals that the model can not understand how to aggregate information from phrase graph into focal phrase embedding without retrieval loss.

\subsection{Supervised Learning with Training Set}


\begin{table}[]

\small
\begin{tabular}{lccc}
\hline Model & Dim. & Pearson cor. & Spearman cor. \\
\hline


$\text{BERT}_{\text{large}}$ & 1024 &  0.704 & 0.683 \\


$\text{RoBERTa}_{\text{large}}$ & 1024 &  0.651 &  0.633 \\

SBERT & 768 & 0.724 & 0.706 \\

\textbf{RA-Sim} & 768 & \textbf{0.741} & \textbf{0.721}
\\
\hline
\end{tabular} \caption{Inference performance in supervised learning.} \label{table:supervised_result}

\end{table}
One may consider using RA-Sim in a supervised learning setting. We conduct supervised learning experiments with the training dataset, and present model performance in Table \ref{table:supervised_result}, with training details in \textbf{Appendix} \ref{app:supervisedsetup}. Generally, the results are consistent with the observations in self-supervised setting: RA-Sim outperforms other benchmarks. Notably, we emphasize that our focus is on the more challenging self-supervised setup with label absence.




\section{Analysis}
We conduct further analyses to understand the inner workings of self-supervised RA-Sim.

\subsection{Alignment and Uniformity}


\begin{figure} 
    \centering
    \includegraphics[width=7.6cm]{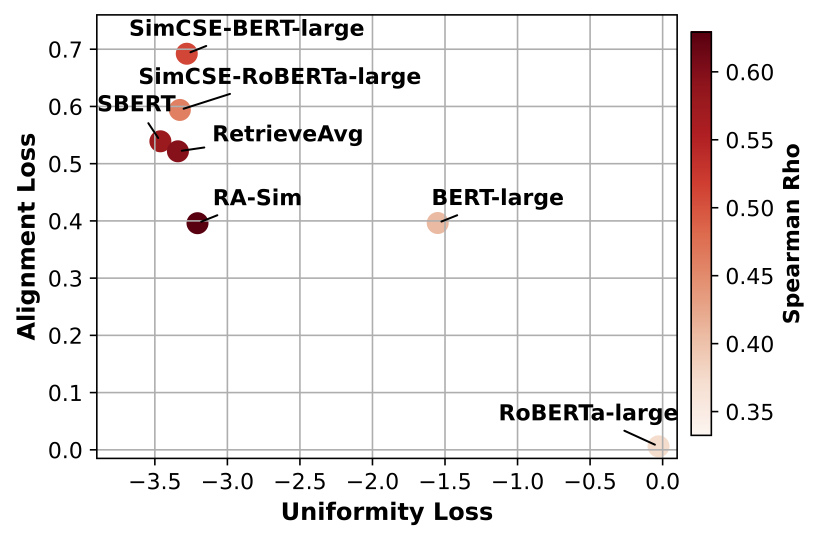}
    \caption{Alignment and Uniformity losses scatter plot. Colors represent similarity inference performance. For clarity, we do not include methods like Word2Vec and Graph-Only in the plot due to poor performance.}
    \label{fig:alignment_uniformity}
\end{figure}

We follow prior work \citep{wang2020understanding} to use alignment and uniformity losses to analyze the underlying mechanism. The a smaller loss in alignment or uniformity implies a better representation capability. Intuitively, a smaller alignment loss indicates that positive pairs are more similar, and a smaller uniformity loss indicates better information retention. We obtain the positive samples in alignment by the similarity threshold of 0.75\footnote{Similar patterns are also observed in other thresholds.}. The scatter plot is shown in Figure \ref{fig:alignment_uniformity}.

Specifically, we observe that: (1) RA-Sim preserves uniformity and enhances alignment, as evidenced by the relative positions between RA-Sim and SBERT; (2) as SimCSE training improves uniformity, it performs poorly in alignment (\textit{i.e.}, inferring positive examples in patent phrases are challenging for SimCSE), which contradicts that SimCSE presents good alignment in general text; (3) a simple retrieval scheme in RetrieveAvg enhances alignment; (4) retrieving the graph yields better alignment than simply retrieving related patents.



\subsection{Qualitative Results}
\begin{table}[]
\small

\begin{tabularx}{\linewidth}{@{}lX|X@{}}
\hline
         & SBERT      & RA-Sim      \\ \hline
\multicolumn{3}{l}{\textbf{Query Phrase}: planting tree} \\ \hline
\#1      &      artificial tree   &   artificial tree       \\ \hline
\#2      &    tilt of tree   &   plant growth       \\ \hline
\#3      &   jackfruit tree   &   man made plant  \\ \hline
\#4      &   tree topology  &   plant  \\ \hline
\#5      &  disjoint or tree  &   plant species  \\ \hline

\multicolumn{3}{l}{\textbf{Query Phrase}: eye quadrants}              \\ \hline
\#1      &  quadrants  &  eye region anatomy \\ \hline
\#2      & eye region anatomy &  rotate eyes  \\ \hline
\#3      &  quadrant   & eye position plan \\ \hline
\#4      &  hand quadrant &  eye image  \\ \hline
\#5      &  shape quadrant  &  apertures of sclera \\ \hline

\end{tabularx} \caption{Retrieved top-5 phrases from testing set.} \label{table:retreival}
\end{table}
Using cosine similarity and trained embeddings, we randomly select phrases as queries to retrieve similar phrases from the testing data. Specifically, we validate RA-Sim from different perspectives and show some results in Table \ref{table:retreival}. While "planting tree" is a commonly used in daily language, "eye quadrants" is a highly technical term. When retrieving "planting tree", the top-5 results in RA-Sim include "artificial tree" and "plant growth", which directly relevant to the query phrase. As for SBERT, the results are noisy with graph topology terms like "tree topology". As for "eye quadrants", both methods have an overlap of "eye region anatomy," which is directly correlated to the query. However, RA-Sim includes more precise results, such as "rotate eyes" and "eye position plan", which are specifically related to the anatomy, positioning, and images of the eye. SBERT includes results like "hand quadrant", which may be less directly related to the eye region anatomy. These results demonstrate the effectiveness of RA-Sim embeddings.

\section{Conclusion}

We explore the challenge of inferring patent phrase similarity, a task distinct from classic semantic textual similarity due to the highly technical language inherent to the patent domain. In this study, we introduce a retrieval-augmented phrase similarity method, termed RA-Sim. This method enhances the contextualized textual information of a patent phrase by incorporating a phrase graph that encompasses its focal patent, related patents via the citation network, and associated phrases through these related patents. RA-Sim is trained using self-supervised learning objectives centered on the phrase graph. Experimental results demonstrate RA-Sim's superior performance, highlighting the benefits of retrieval augmentation for contextualized embeddings in domain problems.

\section{Limitations}

Our method has several limitations to improve in
the future. Firstly, we only test our approach on patent data, and further research is needed to evaluate its potential for scaling up to other application domains, such as scientific articles. Secondly, the computational cost of generating a phrase set for the vast patent universe is high, which is why we utilized the efficient RAKE method. In future work, we plan to explore more computationally expensive phrase generation methods. Lastly, compared to non-graph-based methods, our approach incurs additional computational costs in learning from the retrieved graph module.

\bibliography{paper}

\clearpage
\appendix

\section{Token Statistics}\label{APP:tokenstatistics}
We summarize the token statistics in the following Table \ref{text_statistics}.

\begin{table}[h]

\small

\begin{tabular}{lcccc}
\hline
                & Max  &Min & Average & Std. \\ \hline
Phrase by RAKE  &  11   &  1   &  3.3  &  0.86 \\ 
Patent Abstract & 9,134 & 1 & 114.9  & 49.64 \\ \hline
\end{tabular}
\caption{Statistics of token number.}\label{text_statistics}
\end{table}

\section{Full Training Details} \label{app:fulltrainingdetails}
We train the model with a learning rate from $\{2e^{-6}, 2e^{-5}, 2e^{-4}\}$, a batch size from $\{2, 4, 6, 8\}$, and a maximum of 2 epochs. We experiment with graph layers from 1 to 3 and find that 2 layers performed the best. Our model is trained on 4 GeForce RTX 3090 GPUs, each with 24G memory, using Pytorch\footnote{https://pytorch.org} and PyTorch Geometric\footnote{https://pytorch-geometric.readthedocs.io} for implementation. We use the Python library Pyserini\footnote{https://github.com/castorini/pyserini} to implement BM25 with different retrieval sizes of $k$ from $\{3, 5, 7, 50\}$. For neighbor sampling, we experiment our method with sampling iteration times from $\{1, 2, 3\}$ and per-iteration sampled neighbors from $\{1, 3, 5\}$. Margin parameter $\delta_r$  and $\delta_c$ are chosen from $\{0.01, 0.02, 0.05, 0.1, 0.2, 0.5\}$. Adam \citep{DBLP:journals/corr/KingmaB14} is used to optimize the model parameters in a end-to-end fashion. We assess the model's performance every 100 training steps by evaluating it on the validation set. We choose the best checkpoint according to validation performance and finally evaluate model on testing phrase set. We use Sentence-BERT model with all-mpnet-base-v2\footnote{We examine model performance under different textual encoders like GTE \citep{li2023towards} and E5 \citep{wang2022text}, and all-mpnet-base-v2 performs the best.} as the base textual encoder. The reported results are averaged over 3 runs.

\section{Supervised Learning Setup} \label{app:supervisedsetup}
We conduct supervised learning experiments on BERT, RoBERTa, and SBERT to compare with our method. We obtain embeddings for two phrases and use Mean Squared Error loss to guide training, following prior work \citep{reimers-gurevych-2019-sentence}. We compute the loss for a phrase pair $(u_1, u_2)$ sampled from the training dataset:

\begin{equation} \label{eq:loss_supervised}
\mathcal{L}_{supervised}(u_1, u_2)={MSE}(\hat{y}, y),
\end{equation} 
where $\hat{y} = sim(\phi(u_1), \phi(u_2))$ is inferred similarity for phrases $u_1$ and $u_2$, and $y$ is ground truth similarity labelled by domain experts. As for our method, we train RA-Sim jointly with supervised loss Eq. \ref{eq:loss_supervised} and proposed self-supervised loss Eq. \ref{eq:final_loss} after training with supervised loss Eq. \ref{eq:loss_supervised} solely for 2 epochs. As for other benchmarks, we solely use Eq. \ref{eq:loss_supervised} for training. We limit overall training budget to 5 epochs and show model performance in Table \ref{table:supervised_result}.

\section{Baseline Models} \label{app:baselinemodels}

We elaborate on how we implement different baselines
for comparison in our evaluation:

\textbf{{Pretrained Model}}. Existing pretrained language model can effectively map phrases to embeddings with parameters pretrained on large amounts of training data. Bidirectional Encoder Representations from Transformers (\textbf{BERT}) \citep{devlin-etal-2019-bert} is a widely-adopted pretrained language model and we use mean pooling to obtain phrase-level embeddings. Sentence-BERT \citep{reimers-gurevych-2019-sentence} \textbf{(SBERT)} improves BERT by incorpating dual-tower architecture to obtain sentence embeddings, which is a competitive method in phrase similarity. We use all-mpnet-base-v2 as the base model for Sentence-BERT. \textbf{Patent-BERT} \citep{srebrovic2020leveraging} trains BERT on patent data to obtain a pretrained model on patent. \textbf{Contriever} \citep{izacard2022unsupervised} is trained for unsupervised dense information retrieval and we obtain phrase embeddings by mean pooling on hidden states. GTE utilizes a multi-stage contrastive learning method during the pretraining phase \citep{li2023towards}, while instructor-xl \citep{su2022one} employs instruction to fine-tune and enhance embeddings. Abovementioned pretrained models are fetched from Hugging Face\footnote{https://huggingface.co/models}.

\textbf{{Finetune}}. Simple Contrastive Learning of Sentence Embeddings (SimCSE) finetunes BERT \citep{devlin-etal-2019-bert} and RoBERTa \citep{liu2019roberta} by leveraging an unsupervised loss which views two different embeddings of a same phrase as positive pair in a contrastive objective. We apply SimCSE training to finetune BERT, RoBERTa, SBERT on the mean-pooled hidden states, with the same training phrases as our method. We experiment bert-base-uncased, bert-large-uncased for BERT, termed \textbf{$\text{SimCSE-BERT}_\text{base}$}, \textbf{$\text{SimCSE-BERT}_\text{large}$}. As for RoBERTa, we finetune base models of roberta-base and roberta-large, termed \textbf{$\text{SimCSE-RoBERTa}_\text{base}$}, \textbf{$\text{SimCSE-RoBERTa}_\text{large}$}. The all-mpnet-base-v2 is used for training SBERT based SimCSE, termed \textbf{SimCSE-SBERT}. All aforementioned base models for training SimCSE are from Hugging Face.

\textbf{{Word Embedding}}. \textbf{Glove} \citep{pennington2014glove}, \textbf{Word2Vec} \citep{mikolov2013efficient} and \textbf{FastText} \citep{bojanowski2017enriching} are established off the shelf word-level embeddings. For GloVe we use the Wikipedia 2014 Gigaword 5 model, for FastText the wiki-news-300d-1M model, and for Word2Vec the Wiki-words-250 from TensorFlow Hub\footnote{https://tfhub.dev/google/Wiki-words-250/2}.

\textbf{{Retrieval-based Embedding}}. We propose a retrieval-based baseline, named \textbf{RetrieveAvg}. RetrieveAvg retrieves the most relevant patent for a given phrase using BM25, and then obtains augmented embedding by weighted averaging phrase embedding and most relevant patent's embedding. Specifically, we use SBERT to map phrases or most relevant patents to embeddings. We adjust the weight (a scalar) used to combine the two embeddings according to the performance on validation dataset.

\textbf{{Graph-based Embedding}}. We propose \textbf{Graph-Only}, a graph-based baseline to validate the power of graph structure. This method is implemented by simply replacing the initial node embeddings in RA-Sim with random embeddings to remove the effects of phrase and patent text information.

\section{Examples of Phrase and BM25 Retrieval Results} \label{app:examplesofphraseandbm25}

\subsection{Phrase Examples}\label{appendix_phrase}
We show phrase examples generated by the Rapid Automatic Keyword Extraction (RAKE), shown in Table \ref{table:sample_patent_phrases}.

\begin{table}[th]
\small
\begin{tabularx}{\linewidth}{X|p{1.5cm}}
\hline
Abstract & Phrases \\
\hline 
A multi-layered optical disk comprising a plurality of recording layers accumulated in the thickness direction wherein a light beam is focused on one of tracks of one of the layers thereby to record and reproduce data, the optical disk being characterized in that recording layers each have an identification section storing an address of the recording layer which the identification section belongs to. & Optical disk \par \hspace*{\fill} \par Recording layer \par \hspace*{\fill} \par Light Beam \\

\hline

Provided is a light-emitting element with a small degree of luminance degradation with accumulation of driving time (a long-lifetime light-emitting element). Provided is a light-emitting element in which a light-emitting layer with an electron-transport property is formed with a plurality of layers containing different host materials. Further, the LUMO level of a host material on an anode side is higher than the LUMO level of a host material on a cathode side. With such a structure, it is possible to provide a long-lifetime light-emitting element with little degradation in luminance with accumulation of driving time. & Emitting Element \par \hspace*{\fill} \par Emitting Layer \par \hspace*{\fill} \par Luminance \\ \hline

A computer includes a first memory, a second memory having an I/O speed lower than an I/O speed of the first memory, a storage device, and a processor. The first memory has a work area and a first cache area where data input to and output from the storage device is temporarily stored and the second memory has a second cache area where the data input to and output from the storage device is temporarily stored and a swap area to be a saving destination of data stored in the work area. The processor reduces the work area and expands the first cache area, when an input/output amount to be an amount of data input to and output from the storage device is larger than a predetermined input/output amount. & Work Area \par \hspace*{\fill} \par Second Memory \par \hspace*{\fill} \par Storage Device \\

\hline
\end{tabularx}
\caption{Patent abstract and phrase examples.}
\label{table:sample_patent_phrases}
\end{table}

\subsection{BM25 Retrieval Example}\label{appendix_retrieval}

We present retrieval results for BM25, which retrieves relevant patents for a given phrase from the USPTO patent database. The sample results are shown in Table \ref{table:bm25_results}.

\begin{table}[ht]

\small
\begin{tabularx}{\linewidth}{X}

\hline
\textbf{Query Phrase}: Wireless Communications Network \\ \hline
A method for selecting an alternate wireless communication system for a wireless communication device is disclosed. The method comprises using a first radio access technology (RAT) by a wireless communication device when scanning (202) for an initial wireless communication network (201). The wireless communication device further registers (215) to the initial wireless communication network (201). Then, the initial wireless communication network (201) determines (206) alternate RAT wireless communication network information for the wireless communication device and sends (235) the alternate RAT wireless communication network information to the wireless communication device. The wireless communication device receives the alternate RAT wireless communication network information and selects (240) a first alternate wireless communication network from within the alternate RAT wireless communication network information, scans for the first alternate wireless communication network using an alternate RAT and registers with the first alternate wireless communication network using the alternate RAT. \\ \hline

A wireless communication device comprises a processing system and a wireless communication transceiver. The processing system is configured to store in a memory system data that associates a geographic identifier, a pseudo network signal of a first wireless communication network, and a wireless communication channel of a second wireless communication network. The wireless communication transceiver is configured to wirelessly exchange first wireless communications with the first wireless communication network. The processing system is configured to, in response to the wireless communication device entering a geographic region associated with the geographic identifier, process the geographic identifier to identify the pseudo network signal of the first wireless communication network and the wireless communication channel of the second wireless communication network. The wireless communication transceiver is configured to wirelessly receive the pseudo network signal from the first wireless communication network, and in response, wirelessly exchange second wireless communications with the second wireless communication network over the wireless communication channel. \\ \hline
According to some embodiments, a method in a wireless device operable in a first wireless communication network and a second wireless communication network comprises receiving, from the first wireless communication network, an identification of network nodes of the second wireless communication network. The network nodes of the second wireless communication network are operable to process traffic for the wireless device. The method further comprises receiving an instruction from the second wireless communication network to move traffic from a first network node of the second wireless communication network to a second network node of the second wireless communication network. The first network node is one of the identified one or more network nodes of the second wireless communication network. The method also comprises determining that an identification of the second network node is not included in the received identification of one or more network nodes of the second wireless communication network. \\ \hline

\end{tabularx} 
\caption{Retrieved patents for phrase "Wireless Communications Network". The three most relevant patents are shown.}
\label{table:bm25_results}
\end{table}

\end{document}